\documentclass[letterpaper]{article} % DO NOT CHANGE THIS
\usepackage{aaai24}  % DO NOT CHANGE THIS
\usepackage{times}  % DO NOT CHANGE THIS
\usepackage{helvet}  % DO NOT CHANGE THIS
\usepackage{courier}  % DO NOT CHANGE THIS
\usepackage[hyphens]{url}  % DO NOT CHANGE THIS
\usepackage{graphicx} % DO NOT CHANGE THIS
\usepackage{natbib}  % DO NOT CHANGE THIS AND DO NOT ADD ANY OPTIONS TO IT
\usepackage{caption} % DO NOT CHANGE THIS AND DO NOT ADD ANY OPTIONS TO IT
\usepackage{algorithm}
\usepackage{algorithmic}

\usepackage{newfloat}
\usepackage{listings}
\DeclareCaptionStyle{ruled}{labelfont=normalfont,labelsep=colon,strut=off} % DO NOT CHANGE THIS
\floatstyle{ruled}
\newfloat{listing}{tb}{lst}{}
\floatname{listing}{Listing}
\usepackage{amsmath}
\usepackage{amsfonts}
\usepackage{pgf}
\usepackage{xcolor}

\newcommand{\States}{\mathcal{S}}
\newcommand{\Actions}{\mathcal{A}}
\newcommand{\Goals}{\mathcal{G}}
\newcommand{\augGoals}{\bar{\mathcal{G}}}

\newcommand{\mparams}{{\boldsymbol{\theta}}}
\newcommand{\subroutine}[1]{\texttt{#1}}
\newcommand{\qparams}{{\mathbf{w}}}

\newcommand{\vsg}{r_\gamma}
\newcommand{\vgg}{\tilde{r}_\gamma}
\newcommand{\Gammasg}{\Gamma}
\newcommand{\Gammagg}{\tilde{\Gamma}}
\newcommand{\indsg}{m}
\newcommand{\relsg}{d}

\newcommand{\vsub}{v_{g^\star}}

\newcommand{\rparams}{{\boldsymbol{\theta}^r}}
\newcommand{\gamparams}{{\boldsymbol{\theta}^\Gamma}}
\newcommand{\vggparams}{{\tilde{\boldsymbol{\theta}}^r}}
\newcommand{\gamggparams}{{\tilde{\boldsymbol{\theta}}^\Gamma}}
\newcommand{\vgoal}{\tilde{v}}

\newcommand{\polparams}{{\boldsymbol{\theta}^\pi}}
\newcommand{\rgamma}{r_{\gamma}}
\newcommand{\GamModel}{\Gamma}

\newcommand{\optionq}{\tilde{q}}
\newcommand{\qtarg}{\qparams_{\mathrm{targ}}}

\newcommand{\argmax}{\mathop{\mathrm{arg max}}}
\newtheorem{hypothesis}{\textbf{Hypothesis}}

\title{A New View on Planning in Online Reinforcement Learning}
\author{
    Kevin Roice\equalcontrib \textsuperscript{\rm 1},
    Parham Mohammad Panahi\equalcontrib \textsuperscript{\rm 1},
    Scott M. Jordan\textsuperscript{\rm 1}, \\
    Adam White\textsuperscript{\rm 1, 2 \textdagger},
    Martha White\textsuperscript{\rm 1, 2 \textdagger}
}
\affiliations{
    \textsuperscript{\rm 1}Department of Computing Science, University of Alberta\\
    \textsuperscript{\rm 2}Alberta Machine Intelligence Institute (Amii)\\
    \textsuperscript{\textdagger}Canada CIFAR AI Chair\\
    \{roice, parham1, sjordan, amw8, whitem\}@ualberta.ca
}

\usepackage{bibentry}
\begin{document}

\maketitle

\begin{abstract}
This paper investigates a new approach to model-based reinforcement learning using background planning: mixing (approximate) dynamic programming updates and model-free updates, similar to the Dyna architecture. Background planning with learned models is often worse than model-free alternatives, such as Double DQN, even though the former uses significantly more memory and computation. The fundamental problem is that learned models can be inaccurate and often generate invalid states, especially when iterated many steps. In this paper, we avoid this limitation by constraining background planning to a set of (abstract) subgoals and learning only local, subgoal-conditioned models. This goal-space planning (GSP) approach is more computationally efficient, naturally incorporates temporal abstraction for faster long-horizon planning and avoids learning the transition dynamics entirely. We show that our GSP algorithm can propagate value from an abstract space in a manner that helps a variety of base learners learn significantly faster in different domains.
\end{abstract}

\section{Introduction}

Planning with learned models in reinforcement learning (RL) is important for sample efficiency. Planning provides a mechanism for the agent to generate hypothetical experience, in the background during interaction, to improve value estimates. This hypothetical experience provides a stand-in for the real-world; the agent can generate many experiences (transitions) in its head (using its model) and learn from those experiences. Dyna \citep{sutton1991integrated} is a classic example of this {\em background planning}. On each step, the agent generates several transitions according to its model, and updates its policy with those transitions as if they were real experience. 

Background planning can be used to both adapt to the non-stationarity and exploit aspects of the world that remain constant. In many interesting environments, like the real-world or multi-agent games, the agent will be under-parameterized and thus cannot learn or even represent a stationary optimal policy. The agent can overcome this limitation, however, by using a model to rapidly update its policy. Continually updating the model and replanning allows the agent to adapt to the current situation. In addition, many aspects of the environment remain stationary (e.g., fire hurts and objects fall). The model can capture these stationary facts about how the world works and planning can be used to reason about how the world works to produce a better policy.

The promise of background planning is to learn and adapt value estimates efficiently, but many open problems remain to make it more widely useful. These include that 1) long rollouts generated by one-step models can diverge or generate invalid states, 2) learning probabilities over outcome states can be complex, especially for high-dimensional tasks and 3) planning itself can be computationally expensive for large state spaces. 

One way to overcome these issues is to construct an abstract model of the environment and plan at a higher level of abstraction. In this paper, we construct abstract MDPs using both state abstraction as well as temporal abstraction. State abstraction is achieved by simply grouping states. Temporal abstraction is achieved using \emph{options}---a policy coupled with a termination condition and initiation set \citep{sutton1999options}. A temporally-abstract model based on options allows the agent to jump between abstract states potentially alleviating the need to generate long rollouts.

An abstract model can be used to directly optimize a policy in the abstract MDP, but there are issues with this approach. This idea was explored with an algorithm called Landmark-based Approximate Value Iteration (LAVI) \citep{mann2015approximate}. Though planning can be shown to be provably more efficient, the resulting policy is suboptimal, restricted to going between landmark states. This suboptimality issue forces a trade-off between increasing the size of the abstract MDP (to increase the policy's expressivity) and increasing the computational cost to update the value function. In this paper, we investigate abstract model-based planning methods that have a small computational cost, can quickly propagate changes in value over the entire state space, and do not limit the optimality of learned policy. 

An alternative strategy that we explore in this work is to use the policy optimized in the abstract MDP to guide the learning process in the original MDP. More specifically, the role of the abstract MDP is to propagate value quickly over an abstract state space and then transfer that information to a value function estimate in the original MDP. This has two main benefits: 1) the abstract MDP can quickly propagate value with a small computational cost, and 2) the learned policy is not limited to the abstract value function's approximation. Overall, this approach increases the agent's ability to learn and adapt to changes in the environment quickly. 

Specifically, we introduce Goal-Space Planning (GSP), a new background planning formalism for the general online RL setting. 
The key novelty is designing the framework to leverage \emph{subgoal-conditioned models}: temporally-extended models that condition on subgoals. These models output predictions of accumulated rewards and discounts for state-subgoal pairs, which can be estimated using standard value-function learning algorithms.  The models are designed to be simple to learn, as they are only learned for states local to subgoals and they avoid generating entire next state vectors. We use background planning on transitions between subgoals, to quickly propagate (suboptimal) value estimates for subgoals. We then leverage these quickly computed subgoal values, without suffering from suboptimality, by incorporating them into any standard value-based algorithm via potential-based shaping. In fact, we layer GSP algorithm onto two different algorithms---Sarsa($\lambda$) and Double Deep Q-Network (DDQN)---and show that improves on both base learners.

We carefully investigate the components of GSP, particularly showing that 1) it propagates value and learns an optimal policy faster than its base learner, 2) can perform well with somewhat suboptimal subgoal selection, but can harm performance if subgoals are very poorly selected.
\section{Problem Formulation}
\label{sec:problem_and_bg_planning}

We consider the standard reinforcement learning setting, where an agent learns to make decisions through interaction with an environment, formulated as a Markov Decision Process (MDP) represented by the tuple $(\mathcal{S}, \mathcal{A}, \mathcal{R}, \mathcal{P})$, where $\mathcal{S}$ is the state space and $\mathcal{A}$ is the action space. The reward function $\mathcal{R}: \mathcal{S}\times \mathcal{A} \times \mathcal{S} \rightarrow \mathbb{R}$ and the transition probability $\mathcal{P}: \mathcal{S}\times \mathcal{A} \times \mathcal{S}\rightarrow [0,1]$ describe the expected reward and probability of transitioning to a state, for a given state and action. On each discrete timestep $t$ the agent selects an action $A_t$ in state $S_t$, the environment transitions to a new state $S_{t+1}$ and emits a scalar reward $R_{t+1}$.

The agent's objective is to find a policy $\pi: \States \times \Actions \rightarrow [0, 1]$ that maximizes expected \emph{return}, the future discounted reward  
$G_t \doteq R_{t+1} + \gamma_{t+1} G_{t+1}$ across all states. The state-based discount $\gamma_{t+1} \in [0,1]$ depends on $S_{t+1}$ \citep{sutton2011horde}, which allows us to specify termination. If $S_{t+1}$ is a terminal state, then $\gamma_{t+1} = 0$; else, $\gamma_{t+1} = \gamma_c$ for some constant $\gamma_c \in [0,1]$.
%on the transition $(S_t, A_t, S_{t+1})$~\citep{white2017unifying}, 
%allowing for a unified treatment of episodic and continuing problems. For continuing problems, the discount may simply be a constant less than 1. For episodic problems the discount might be 1 during the episode, and become zero when $S_t, A_t$ leads to termination. 
The policy can be learned using algorithms like Sarsa($\lambda$) \citep{sutton2018reinforcement}, which approximate the action-values: the expected return from a given state and action, $q(s,a) \doteq \mathbb{E}\left [ G_t | S_t =s, A_t=a \right ]$.

Most model-based reinforcement learning systems learn a state-to-state transition model. 
The transition dynamics model can be either an expectation model $\mathbb{E}[S_{t+1} | S_{t}, A_{t}]$ or a probabilistic model $P(S_{t+1} | S_{t}, A_{t})$.
If the state space or feature space is large, then the expected next state or distribution over it can be difficult to estimate, as has been repeatedly shown \citep{talvitie2017selfcorrecting}. Further, these errors can compound when iterating the model forward or backward \citep{jafferjee2020hallucinating,vanhasselt2019when}. It is common to use an expectation model, but unless the environment is deterministic or we are only learning the values rather than action-values, this model can result in invalid states and detrimental updates \citep{wan2019planning}. The goal in this work is to develop a model-based approach that avoids learning the state-to-state transition model, but still obtains the benefits of model-based learning for faster learning and adaptation.

\begin{figure}[tbp]
     \centering
   \includegraphics[width=0.45\textwidth]{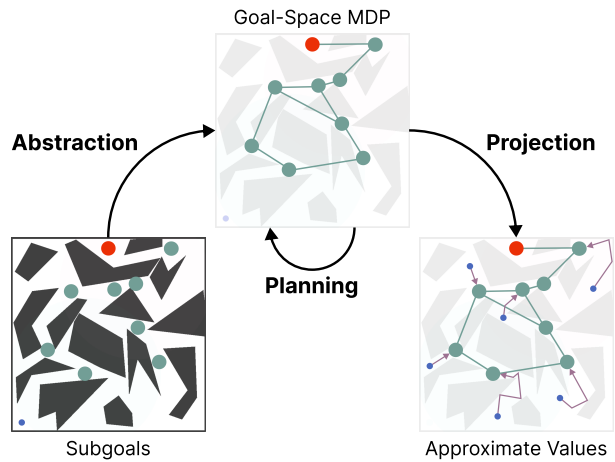}
   \caption{GSP in the PinBall domain. The agent begins with a set of subgoals (denoted in teal) and learns a set of subgoal-conditioned models. \textbf{(Abstraction)} Using these models, the agent forms an abstract MDP where the states are subgoals with options to reach each subgoal as actions. \textbf{(Planning)} The agent plans in this abstract MDP to quickly learn the values of these subgoals. \textbf{(Projection)} Using learned subgoal values, the agent obtains approximate values of states based on nearby subgoals and their values. These quickly updated approximate values are then used to speed up learning. }\label{fig:gsp_overview_diag}  
 \end{figure}

\section{Goal Space Planning}
We consider three desiderata for when a model-based approach should be effective. (1) The model should be feasible to learn: we can get it to a sufficient level of accuracy that makes it beneficial to  plan with that model. (2) Planning should be computationally efficient, so that the agent's values can be quickly updated. (3) Finally, the model should be modular---composed of several local models or those that model a small part of the space---so that the model can quickly adapt to small changes in the environment. These small changes might still result in large changes in the value function; planning can quickly propagate these small changes, potentially changing the value function significantly.

At a high level, the GSP algorithm focuses on planning over a given set of abstract subgoals to provide quickly updated, approximate values to speed up learning. In order to do so, the agent first learns a set of \emph{subgoal-conditioned models}: minimal models focused around planning utility. These models then form a temporally abstract goal-space semi-MDP, with subgoals as states, and options to reach each subgoal as actions. Finally, the agent can update its policy based on these subgoal values to speed up learning. 

Figure \ref{fig:gsp_overview_diag} provides a visual overview of this process. We visualize this is an environment called PinBall, which we also use in our experiments. PinBall is a continuous state domain where the agent must navigate a ball through a set of obstacles to reach the main goal, with a four-dimensional state space consisting of $(x,y, \dot{x}, \dot{y})$ positions and velocities. In this figure, the set of subgoals $\mathcal{G}$ are the teal dots, a finite space of 9 subgoals. The subgoals are abstract states, in that they correspond to many states: a subgoal is any $(x,y)$ location in a small ball, at any velocity. In this example, the subgoals are randomly distributed across the space. Subgoal discovery---identifying this set of subgoals $\mathcal{G}$---is an important pre-requisite for this algorithm. For this paper, however, we focus on this planning formalism assuming these subgoals are given, already discovered by the agent.

In the planning step (top central image in Figure \ref{fig:gsp_overview_diag}), we treat $\mathcal{G}$ as our finite set of states and do value iteration. The precise formula for this update is described in \citeauthor{lo2024goalspace} \shortcite{lo2024goalspace}, along with the formal definition of the models that we learn for goal-space planning. In words, we compute the subgoal values $\vgoal: \mathcal{G} \rightarrow \mathbb{R}$, using $\vgg(g,g') = $ discounted return when trying to reach $g'$ from $g$ and $\Gammagg(g,g') = $ discounted probability of reaching $g'$ from $g$, 
\begin{equation}
\vgoal(g) = \max\limits_{\substack{\text{relevant/nearby} \\ \text{subgoals } g'}} \vgg(g,g') + \Gammagg(g,g') \vgoal(g'). 
\end{equation}

The projection step involves updating values for states, using the subgoal values. The most straightforward way to obtain the value for a state is to find the nearest subgoal $s$ and reason about
$\vsg(s,g) = $ discounted return when trying to reach $g$ from $s$ and $\Gammasg(s,g) = $ discounted probability of reaching $g$ from $s$, 
\begin{equation}
\label{eq:vsub}
\vsub(s) = \max\limits_{\substack{\text{relevant/nearby} \\ \text{subgoals } g}} \rgamma(s, g) + \GamModel(s,g) \vgoal(g).
\end{equation}
Relevance here is defined as $s$ being within the initiation set of the option that reaches that subgoal. We learn an option policy to reach each subgoal, where the initiation set for the option is the set of states from which the option can be executed. The initiation set is a local region around the subgoal, which is why we say we have many local models. Again, we refer the reader to \citeauthor{lo2024goalspace} for the formal definitions of these value functions.

There are several ways we can use this value estimate: inside an actor-critic architecture or as a bootstrap target. For example, for a transition $(s,a,r,s')$, we could update action-values $q(s,a)$ using $r + \gamma \vsub(s')$. This naive approach, however, can result in significant bias, as found in \citeauthor{lo2024goalspace} Instead, we propose an approach to use $\vsub$ as a potential function for potential-based reward shaping \citep{ng1999reward}. For example, in the Sarsa($\lambda$) algorithm, the update for the weights $\qparams$ of the function $q \colon \mathcal{S} \times \mathcal{A} \times \mathbb{R}^n \to \mathbb{R}$ would use the TD-error
\begin{equation} \label{eq:oci_delta}
\begin{split}
\delta_t :=& R_{t+1} + \gamma_{t+1} \vsub(S_{t+1}) - \vsub(S_t) \, + \\ & \, \gamma_{t+1} q(S_{t+1},A_{t+1}; \qparams) - q(S_t,A_t; \qparams).
\end{split}
\end{equation}

A key part of this algorithm is learning the subgoal models, $\vsg$ and $\Gammasg$. These models will be recognizable to many: they are universal value function approximators (UVFAs) \citep{schaul2015universal}. We can leverage advances in learning UVFAs to improve our model learning. These models are quite different from standard models in RL, in that most models in RL input a state (or abstract state) and action and output an expected next state (or expected next abstract state). Essentially, the model inputs the source and outputs the expected destination, or a distribution over the possible destinations. Here, the models take as inputs both the source 
and destination, and output only scalars (accumulated reward and discounted probability). The design of GSP is built around using these types of models, that avoids outputting predictions about entire state vectors.

\begin{figure}[htbp]
	\centering
	\includegraphics[width=0.35\textwidth]{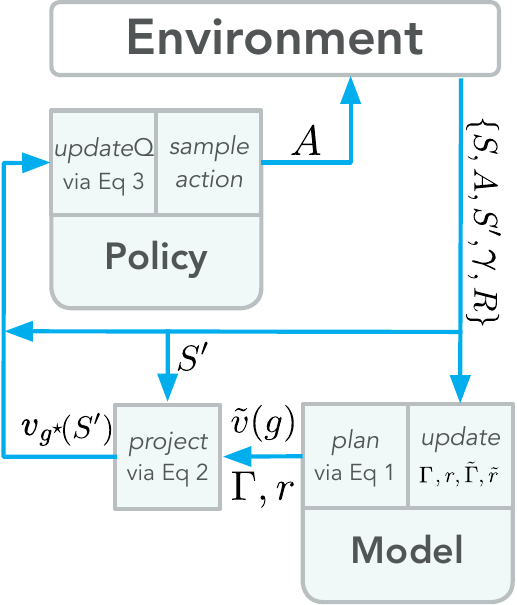}
	\caption{Goal-Space Planning.}\label{gsp_diagram}
\end{figure}

The algorithm is visualized in Figure \ref{gsp_diagram}. The steps of agent-environment interaction include:
\begin{enumerate}
    \item take action $A_t$ in state $S_t$, to get $S_{t+1}, R_{t+1}$ and $\gamma_{t+1}$
    
    \item query the model for $\rgamma(S_{t+1},g)$, $\Gammasg(S_{t+1},g)$, $\vgoal(g)$ for all $g$ where $\relsg(S_{t+1},g) > 0$

    \item compute projection $\vsub(S_{t+1})$, using \eqref{eq:vsub}
    
    \item update the main policy with the transition and $\vsub(S_{t+1})$, using \eqref{eq:oci_delta}. 
\end{enumerate} 
All background computation is used for model learning using a replay buffer and for planning to obtain $\vgoal$, so that they can be queried at any time on step 2.

\begin{figure*}[htbp]
  \centering
   \includegraphics[width=\textwidth]{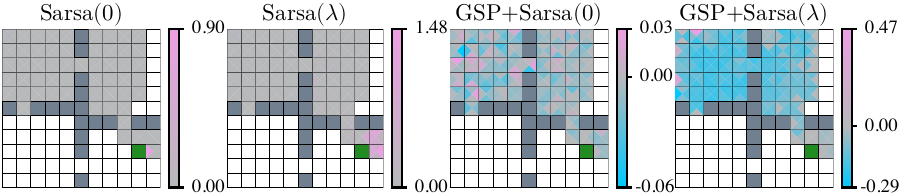}
  \caption{These four plots show the action values after a single episode of updates for Sarsa with and without GSP and eligibility traces, i.e., $\lambda = 0.9$. Each algorithm's update is simulated from the same data collected from a uniform random policy. Each state (square) is made up of four triangles representing each of the four available actions. White squares represent states not visited in the episode.}
  \label{fig:hm_fourrooms}
\end{figure*}

To be more concrete, Algorithm \ref{alg:DDQN_GSP} shows the GSP algorithm, layered on DDQN \citep{van2016deep}. DDQN is a variant of DQN---and so relies on replay---that additionally incorporates the idea behind Double Q-learning to avoid maximization bias in the Q-learning update \citep{hasselt2010double}. All new parts relevant to GSP are colored blue; without the blue components, it is a standard DDQN algorithm. The primary change is the addition of the potential to the action-value weights $\qparams$, with the other blue lines primarily around learning the model and doing planning. GSP should improve on replay because it simply augments replay with a  potential difference that more quickly guides the agent to take promising actions.

\begin{algorithm}[H]
  \caption{Goal-Space Planning for Episodic Problems}
  \label{alg:gsp}
\begin{algorithmic}
\STATE Assume given subgoals $\Goals$ and relevance function $d$
\STATE Initialize base learner (i.e. $\qparams, \mathbf{z} = \mathbf{0}, \mathbf{0}$ for Sarsa($\lambda$)\footnotemark), model parameters $\mparams = (\rparams, \gamparams, \polparams), \tilde{\mparams} = (\vggparams, \gamggparams)$
\STATE Sample initial state $s_0$ from the environment
  \FOR {$t \in 0, 1, 2, ...$}
    \STATE Take action $a_t$ using $q$ (e.g., $\epsilon$-greedy), observe $ r_{t+1}, s_{t+1}, \gamma_{t+1}$
    \STATE Choose $a'$ from $s_{t+1}$ using $q$ (e.g. $\epsilon$-greedy)
    \STATE \subroutine{ModelUpdate}$(s_t, a_t, r_{t+1}, s_{t+1}, \gamma_{t+1})$
    \STATE \subroutine{Planning}$()$
    \STATE \subroutine{MainPolicyUpdate}$(s_t, a_t, s_{t+1}, r_{t+1}, \gamma_{t+1}, a')$
  \ENDFOR
\end{algorithmic}
\end{algorithm}
\footnotetext{Sarsa($\lambda$) has two sets of parameters to initialize: its action-value function weights $\mathbf{w}$, and its eligibility trace vector $\mathbf{z}$ \citep{rummery1995problem}.}

\section{Experiments}

This section motivates the capabilities of the GSP framework through a series of demonstrative results. We investigate the utility of GSP in propagating value and speeding up learning. We do so using learners in three domains: FourRooms, PinBall \citep{konidaris2009skill} and GridBall (a version of PinBall without velocities). Unless otherwise stated, all learning curves are averaged over 30 runs, with shaded regions representing one standard error.

\subsection{GSP on Propagating Value} \label{sec:exp_specification}

The central hypothesis of this work is that GSP can accelerate value propagation. By using information from local models in our updates, our belief is that GSP will have a larger change in value to more states, leading to policy changes over larger regions of the state space.

In this section, we consider the effect of our background planning algorithm on value-based RL methods.
\begin{hypothesis}
GSP changes the value for more states with the same set of experience.
\label{hyp:val_prop_state}
\end{hypothesis}

In order to verify whether GSP helps to quickly propagate value, we first test this hypothesis in a simple grid world environment: the FourRooms domain. The agent can choose from one of 4 actions in a discrete action space $\mathcal{A} = \{\texttt{up}, \texttt{down}, \texttt{left}, \texttt{right}\}$. All state transitions are deterministic. The grey squares in Figure \ref{fourrooms} indicate walls, and the state remains unchanged if the agent takes an action that leads into a wall. This is an episodic task, where the base learner has a fixed start state and must navigate to a fixed goal state where the episode terminates. Episodes can also terminate by timeout after 1000 timesteps.

In this domain, we test the effect of using GSP with pre-trained models on a Sarsa($\lambda$) base learner in the tabular setting (i.e. no function approximation for the value function). Full details on using GSP with this temporal difference (TD) learner can be found in Algorithm \ref{alg:MainPolicySarsaLambdaUpdate}. We set the four hallway states plus the goal state as subgoals, with their initiation sets being the two rooms they connect. Full details of option policy learning can be found in the Appendix \ref{app:opt_pol}.

Figure \ref{fig:hm_fourrooms} shows the base learner's action-value function after a single episode using four different algorithms: Sarsa($0$), Sarsa($\lambda$), Sarsa($0$)+GSP, and Sarsa($\lambda$)+GSP. In Figure \ref{fig:hm_fourrooms}, the Sarsa($0$) learner updates the value of the state-action pair that immediately preceded the +1 reward at the goal state. The plot for Sarsa($\lambda$) shows a decaying trail of updates made at the end of the episode, to assign credit to the state-action pairs that led to the +1 reward. The plots for the GSP variants show that all state-action pairs sampled receive instant feedback on the quality of their actions. The updates with GSP can be both positive or negative based on if the agent makes progress towards the goal state or not. This direction of update comes from the potential-based reward shaping rewards or penalizes transitions based on whether $\gamma_{t+1} \vsub(S_{t+1}) > \vsub(S_{t})$. It is clear that projecting subgoal values from the abstract MDP leads to action-value updates over more of the visited states, even without memory mechanisms such as eligibility traces. 

It is evident from these updates over a single episode that the resulting policy from GSP updates should be more likely to go to the goal. We would like to quantify how much faster this propagated value can help our base learner over multiple episodes of experience. More specifically, we want to test the following hypothesis.
\begin{hypothesis}
GSP enables a TD base-learner to learn faster. 
%than standard TD-based algorithms.
\label{hyp:val_prop_time}
\end{hypothesis}

\begin{figure}[b]
  \centering
    \centering
    % \scalebox{0.55}{\input{figures/figsv2/fourrooms_tab_stepstogoal_noLAVI.pgf}}
    \includegraphics[width=0.4\textwidth]{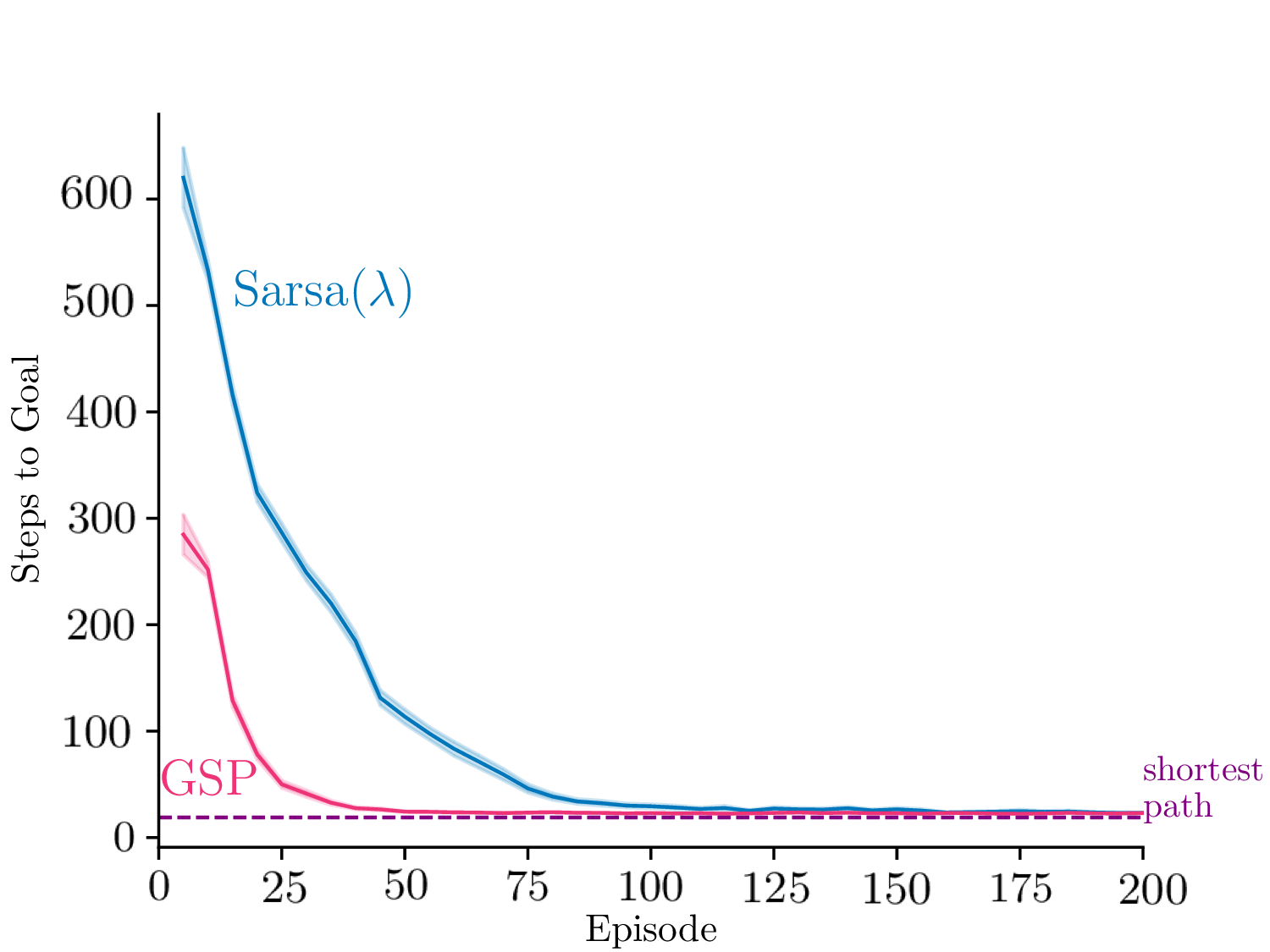}
  \caption{This plot shows the average number of steps to goal smoothed over five episodes in the FourRooms domain. Shaded region represents 1 standard error across 100 runs.}
  \label{fig:fourrooms_curve} 
\end{figure}

We expect GSP to improve a base learner’s performance on a task within fewer environment interactions. We shall test whether the value propagation over the state-action space as seen in Figure \ref{fig:hm_fourrooms} makes this the case over the course of several episodes (i.e. we are now testing the effect of value propagation over time). Figure \ref{fig:fourrooms_curve} shows the performance of a Sarsa($\lambda$) base learner with and without GSP in the FourRooms domain with a reward of -1 per step. Full details on the hyperparameters used can be found in Appendix \ref{app:opt_pol}. It is evident that the GSP-augmented Sarsa($\lambda$) learner is able to reach the optimal policy much faster. The GSP learner also starts at a much lower steps-to-goal.  
We \emph{believe} this first episode performance improvement is because the feedback from GSP teaches the agent which actions move towards or away from the goal during the first episode.

\subsection{GSP in Larger State Spaces}\label{sec:larger_state}
Many real world applications of RL involve large and/or continuous state spaces. Current planning techniques struggle with such state spaces. This motivates an investigation into how well Hypotheses \ref{hyp:val_prop_state} and \ref{hyp:val_prop_time} hold up when GSP is used in such environments (e.g. the PinBall domain). To better analyse GSP and its value propagation across state-space, we also created an intermediate environment between FourRooms and PinBall called GridBall. 

 In all our PinBall experiments, the agent is initialised with zero velocity at a fixed start position at the beginning of every episode. It should be noted that, unlike in the FourRooms environment, there exists states which are not in the initiation set of any subgoal - a common occurence when deploying GSP in the state spaces of real-world applications.

GridBall is like PinBall, but change to be more like a gridworld to facilitate visualization. The velocity components of the state are removed, meaning the state only consists of $(x,y)$ locations, and the action space is changed to displace the ball by a fixed amount in each cardinal dimension. We keep the same obstacle collision mechanics and calculations from PinBall. Since GridBall does not have any velocity components, we can plot heatmaps of value propagation without having to consider the velocity at which the agent arrived at a given position. 

For Hypothesis \ref{hyp:val_prop_state}, we repeat the experiments on GridBall with base learners that use tile-coded value features \citep{sutton2018reinforcement}, and linear value function approximation. Full details on the option policies and subgoal models used for this are outlined in shown in Appendices \ref{app:opt_pol} and \ref{app:model_learning}. Like in the FourRooms experiment, we set the reward to be 0 at all states and +1 once the agent reaches any state in the main goal to show value propagation. We collect a single episode of experience from the Sarsa($0$)+GSP learner and use its trajectory to perform a batch update on all learners. This controls for any variability in trajectories between learners, so we can isolate and study the change in value propagation. 

\begin{figure*}[htbp] % You can adjust the placement options as needed
  \centering
  \begin{minipage}{0.22\textwidth}
    \centering
    \caption*{Sarsa(0)}
    \includegraphics[width=\textwidth]{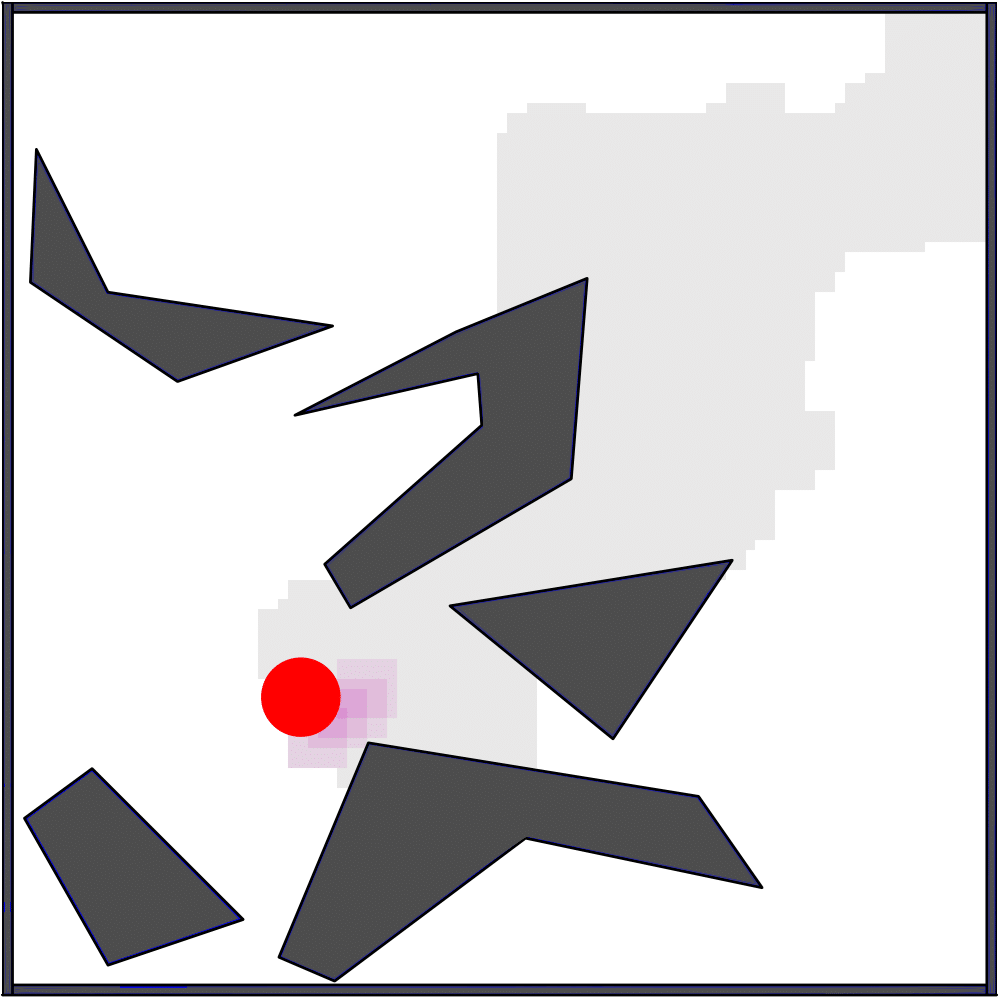}
    \label{sub_fig:hm_gridball_sarsa}
  \end{minipage}\hfill
  \begin{minipage}{0.22\textwidth}
    \centering
    \caption*{Sarsa($\lambda$)}
    \includegraphics[width=\textwidth]{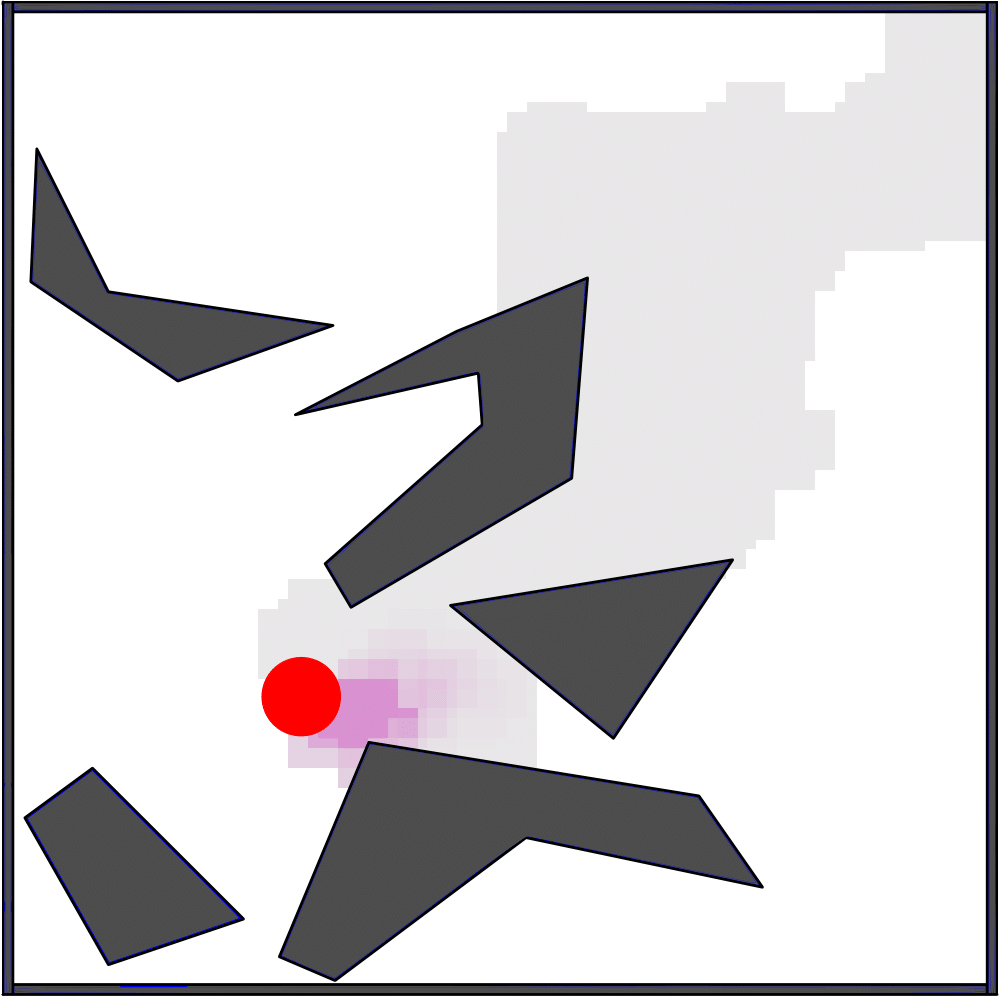}
    \label{sub_fig:hm_gridball_gsp}
  \end{minipage}\hfill
  \begin{minipage}{0.22\textwidth}
    \centering
    \caption*{GSP+Sarsa(0)}
    \includegraphics[width=\textwidth]{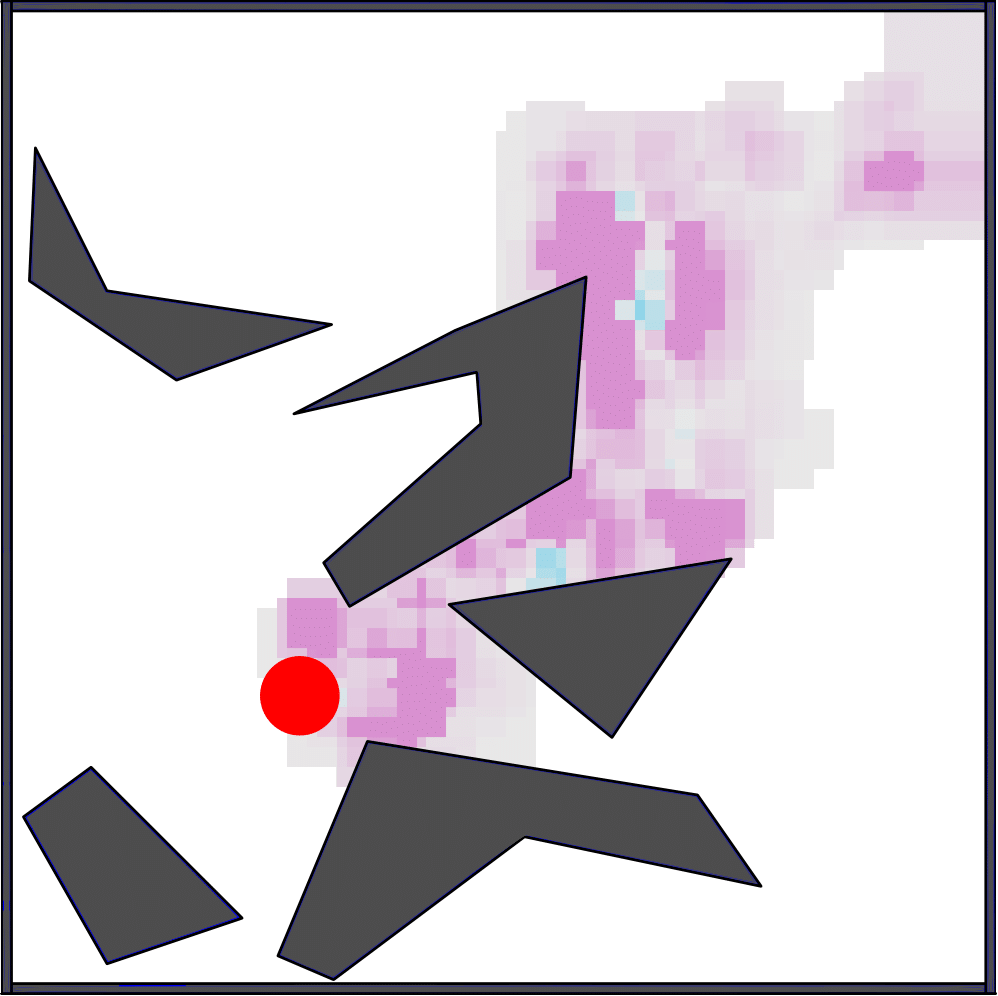}
    \label{sub_fig:hm_gridball_nogsp}
  \end{minipage}\hfill
  \begin{minipage}{0.22\textwidth}
    \centering
    \caption*{GSP+Sarsa($\lambda$)}
    \includegraphics[width=\textwidth]{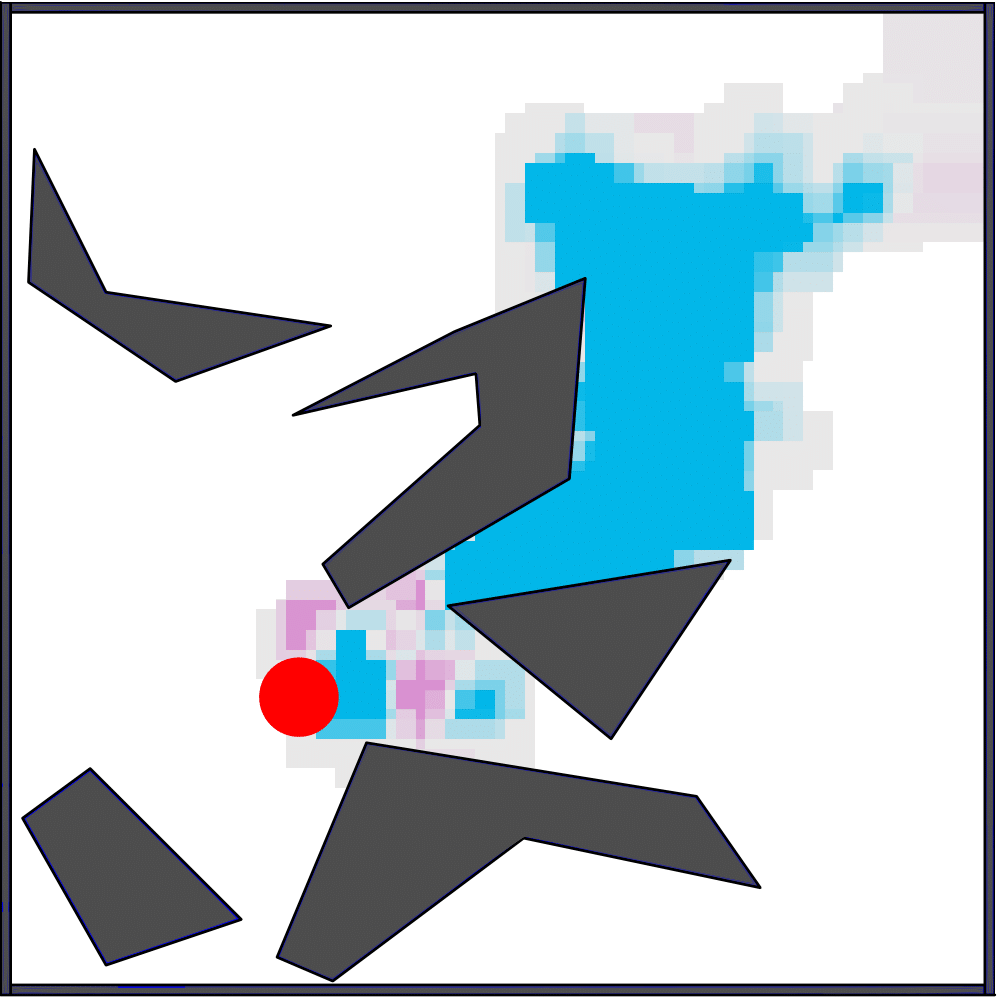}
    \label{sub_fig:hm_gridball_gsp_lambda}
  \end{minipage}\hfill
  \begin{minipage}{0.06\textwidth}
    \includegraphics[width=\textwidth]{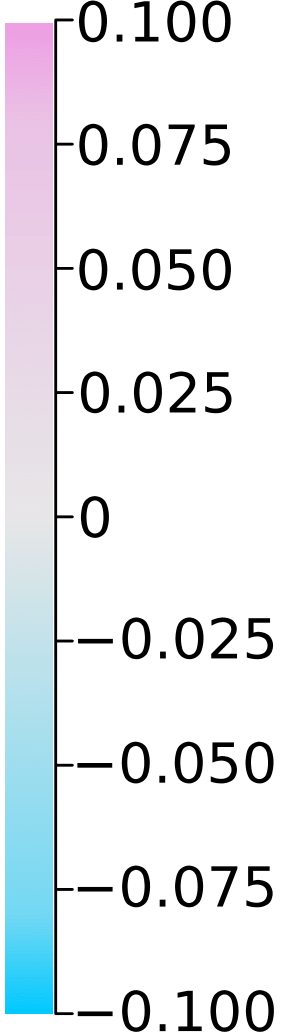}
  \end{minipage}\hfill
  \caption{The tile-coded value function after one episode in GridBall. Like Figure \ref{fig:hm_fourrooms}, the gray regions show the visited states which were not updated. The red circle is the main goal.}
  \label{hm_gridball}
\end{figure*}

Figure \ref{hm_gridball} compares the state value function (averaged over the action value estimates) of Sarsa($0$), Sarsa($\lambda$), Sarsa($0$)+GSP and Sarsa($\lambda$)+GSP learners after a single episode of interaction with the environment. The results are similar to those on FourRooms. The Sarsa($0$) algorithm only updates the value of the tiles activated by the state preceding the goal. Sarsa($\lambda$) has a decaying trail of updates to the tiles activated preceeding the goal, and the GSP learners updates values at all states in the initiation set of a subgoal.

\begin{figure}[tbp]
  \centering
  \begin{minipage}{0.23\textwidth}
    \centering
    \caption*{GridBall}
    % \scalebox{0.4}{\input{figures/figsv2/gridball_stepstogoal_noLAVI.pgf}}
    \includegraphics[width=\textwidth]{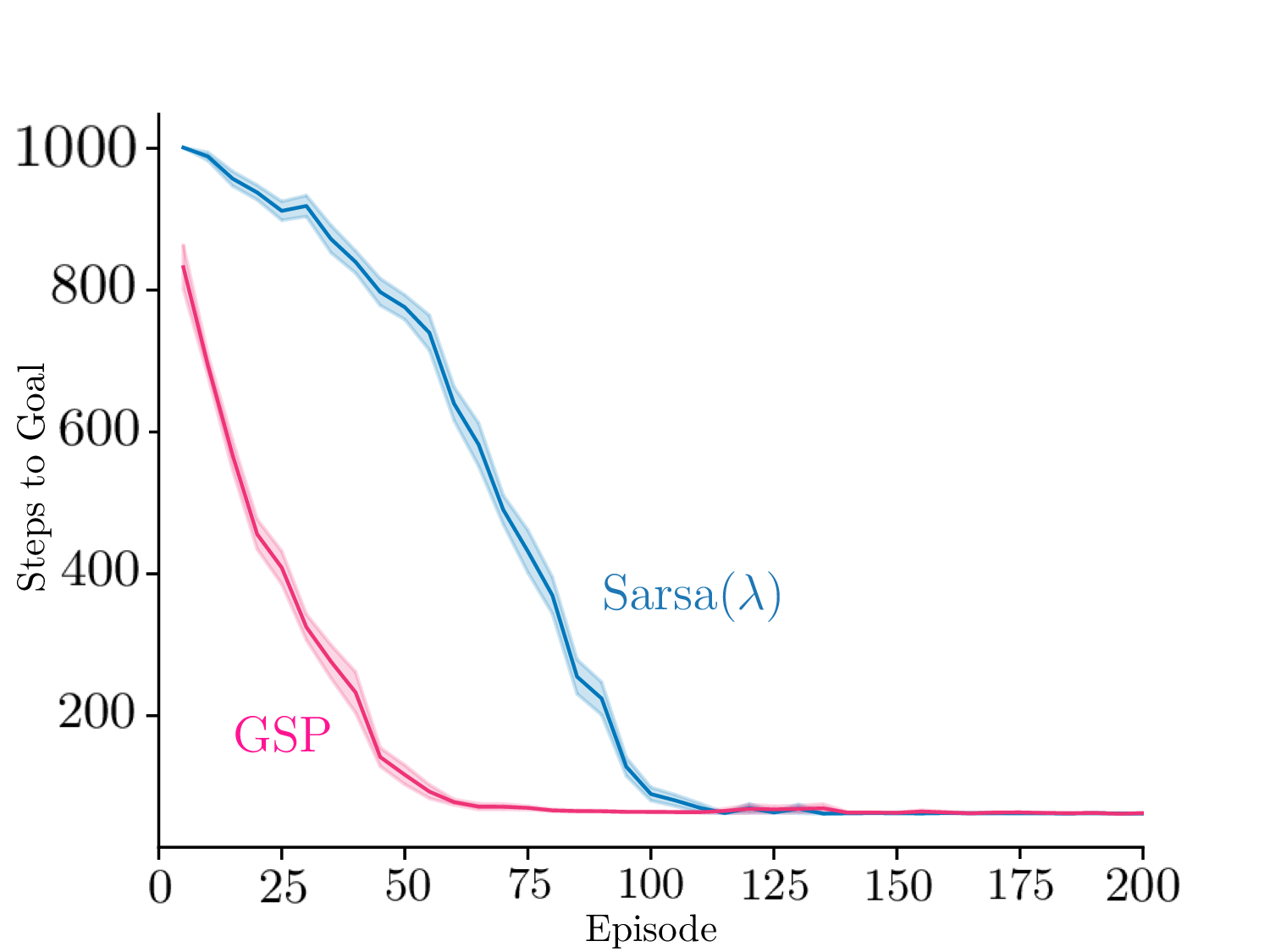}
  \end{minipage}\hfill
  \begin{minipage}{0.23\textwidth}
    \centering
    \caption*{PinBall}
    % \scalebox{0.36}{\input{figures/figsv2/pinball_stepstogoal_noLAVI.pgf}}
    \includegraphics[width=\textwidth]{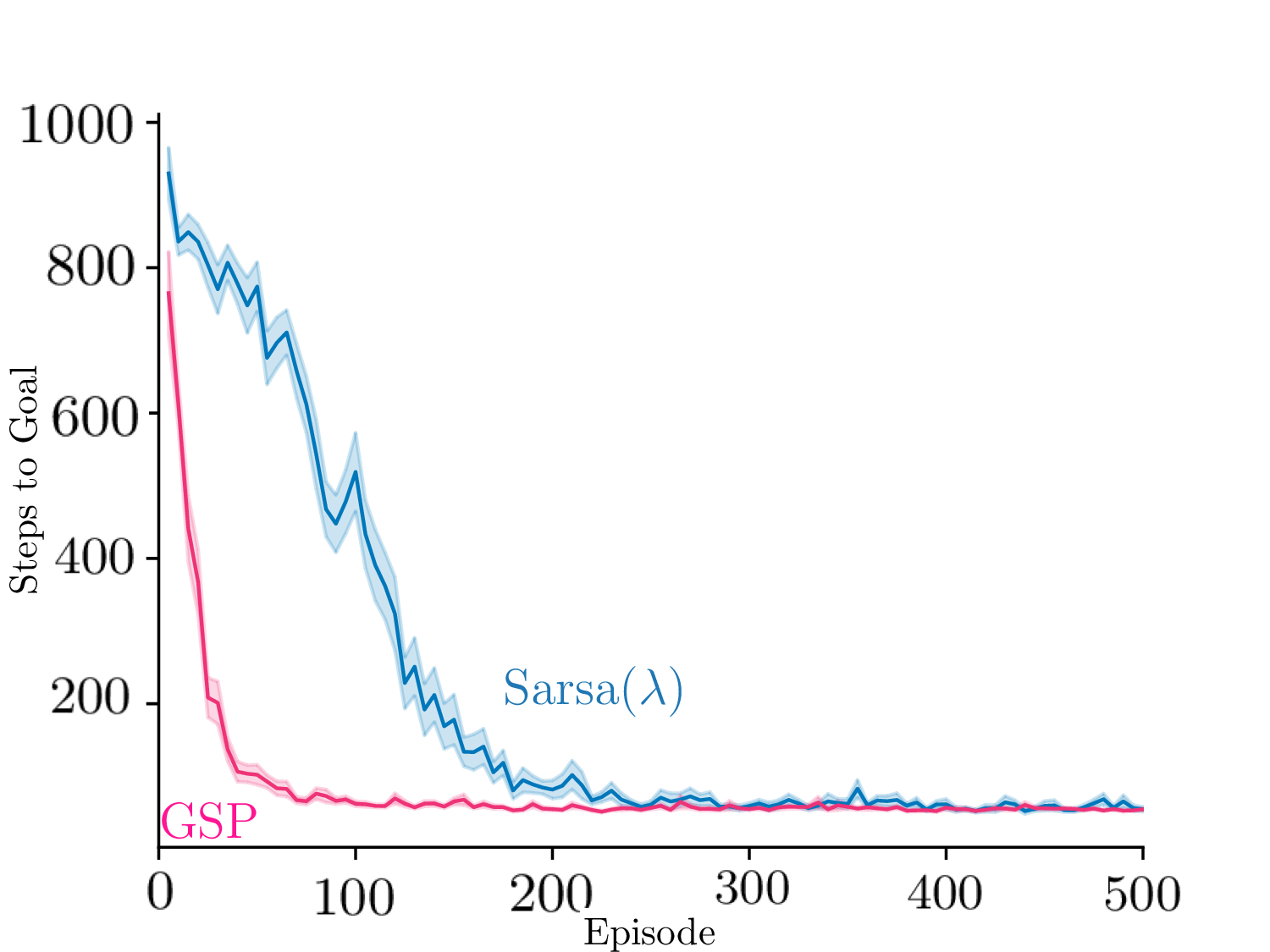}
  \end{minipage}
  \caption{Five episode moving average of return in the GridBall over 200 episodes (left) and PinBall over 500 episodes (right). All learners used linear value function approximation on their tile coded features.}
  \label{fig:gridball_pinball_curve} 
\end{figure}

To examine how GSP translates to faster learning (Hypothesis \ref{hyp:val_prop_time}), we measure the performance (steps to goal) over time for each algorithm in both GridBall and PinBall domains. Figure \ref{fig:gridball_pinball_curve} shows that GSP significantly improves the rate of learning in these larger domains too, with the base learner able to reach near its top performance within 75 and 100 epsiodes in GridBall and PinBall respectively.  All runs are able to find a similar length path to the goal. As the size of the state space increases, the benefit of using local models in the GSP updates still holds. 

Similar to the previous domains, the Sarsa($\lambda$) learner using GSP is able to reach a good policy much faster than the base learner without GSP. In both domains, the GSP and non-GSP Sarsa($\lambda$) learners plateau at the same average steps to goal. Even though the obstacles remain unchanged from GridBall, it takes roughly 50 episodes longer for even the GSP variant to reach a good policy in PinBall. This is likely due to the continuous 4-dimensional state space making the task harder.

\subsection{GSP with Deep Reinforcement Learning}\label{sec:non_linear}
% MARTHA: Too much preamble, I think most people agree we should do deep RL
%Another aspect of using model-based RL in the real world is the ability to generalise across large state spaces. Deep artificial neural networks have shown great success in learning good representations of such state spaces. Deep RL has become commonplace for this - so much so that there have even been recent successes in deep model-based RL \citep{schrittwieser2020mastering, hafner2023mastering}. 
The previous results shed light on the dynamics of value propagation with GSP when a learner is given a representation of it's environment (a look-up table or a tile coding). A natural next step is to look at the whether the reward and transition dynamics learnt in GSP can still propagate value (Hypothesis \ref{hyp:val_prop_time}) in the deep RL setting, where the learner must also learn a representation of its environment.
We test this by running a DDQN base learner \citep{van2016deep} in the PinBall domain, with GSP layered on DDQN as in Algorithm \ref{alg:DDQN_GSP}.

% \begin{figure}[th]
%   \begin{subfigure}{0.49\textwidth}
%     \centering
%     \caption{GSP modified for Deep RL}
%     \scalebox{0.45}{\input{figures/figsv2/ddqn.pgf}}
%   \label{fig:dqn} 
% \end{subfigure}
%   \begin{subfigure}{0.49\textwidth}
%     \centering
%     \caption{Robustness to model accuracy}
%     \scalebox{0.45}{\input{figures/figsv2/partial_fixed.pgf}}
%   \label{fig:partial_mods} 
% \end{subfigure}
%   \caption{Investigating the behavior of GSP in the deep reinforcement learning setting in PinBall. (a) Following the format of Figure \ref{fig:gridball_pinball_curve}, we show the 20 episode moving average of steps to the main goal in PinBall. 
%   (b) Five episode moving average of steps to goal in PinBall for GSP with models trained with differing numbers of epochs.}
%   \label{fig:deep} 
% \end{figure}

\begin{figure}[htbp]
  \centering
  % \begin{minipage}{0.2\textwidth}
    \centering
    % \caption*{GSP modified for Deep RL}
    \scalebox{0.43}{\input{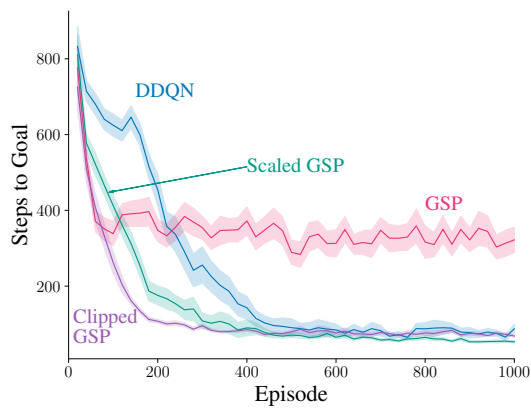}}
  % \end{minipage}\hfill
  % % % \begin{minipage}{0.2\textwidth}
    % % \centering
    % \caption*{Robustness to model accuracy}
    % \scalebox{0.3}{\input{figures/figsv2/partial_fixed.pgf}}
  % \end{minipage}
  \caption{Investigating the behavior of GSP in the deep reinforcement learning setting in PinBall. Following the format of Figure \ref{fig:gridball_pinball_curve}, we show the 20 episode moving average of steps to the main goal in PinBall.}
  \label{fig:deep} 
\end{figure}

% MArtha: contrasting to Sarsa distracts the reader and is not the key point, so omitting
%As expected, DDQN takes more episodes than Sarsa($\lambda$) did to learn a good policy in PinBall. This is because the base learner now also has to learn a competent representation of the state-action space. 

Unlike the previous experiments, using GSP out of the box resulted in the base learner converging to a sub-optimal policy. 
This is despite the fact that we used the same $\vsub$ as the previous PinBall experiments.
We investigated the distribution of shaping terms added to the environment reward and observed that they were occasionally an order of magnitude greater than the environment reward.
Though the linear and tabular methods handled these spikes in potential difference gracefully, these large displacements seemed to causes issues when using neural networks and a DDQN base learner. 

We tested two variants of GSP that better control the magnitudes of the raw potential differences ($\gamma \vsub(S_{t+1}) - \vsub(S_t)$). We adjusted for this by either clipping or down-scaling the potential difference added to the reward. The scaled reward multiplies the potential difference by 0.1. Clipped GSP clips the potential difference into the $[-1, 1]$ interval. With these basic magnitude controls, GSP again learns significantly faster than its base learner, as shown in Figure \ref{fig:deep}.

\section{Related Work}

A variety of approaches have been developed to handle issues with learning and iterating one-step models. Several papers have shown that using forward model simulations can produce simulated states that result in catastrophically misleading values \citep{jafferjee2020hallucinating,vanhasselt2019when,lambert2022investigating}. This problem has been tackled by using reverse models \citep{pan2018organizing,jafferjee2020hallucinating,vanhasselt2019when}; primarily using the model for decision-time planning \citep{vanhasselt2019when,silver2008samplebased,chelu2020forethought}; and improving training strategies to account for accumulated errors in rollouts \citep{talvitie2014model,venkatraman2015improving,talvitie2017selfcorrecting}. 
An emerging trend is to avoid approximating the true transition dynamics, and instead learn dynamics  tailored to predicting values on the next step correctly \citep{farahmand2017valueaware,farahmand2018iterative,ayoub2020modelbased}. This trend is also implicit in the variety of techniques that encode the planning procedure into neural network architectures that can then be trained end-to-end \citep{tamar2016value,silver2017predictron,oh2017value,weber2017imaginationaugmented,farquhar2018treeqn,schrittwieser2020mastering}. We similarly attempt to avoid issues with iterating models, but do so by considering a different type of model.

Current deep model-based RL techniques plan in a lower-dimensional abstract space where the relevant features from the original high-dimensional experience are preserved, often refered to as a \emph{latent space}. MuZero \citep{schrittwieser2020mastering}, for example, embeds the history of observations to then use predictive models of values, policies and one-step rewards. Using these three predictive models in the latent space guides MuZero’s Monte Carlo Tree Search without the need for a perfect simulator of the environment. Most recently, DreamerV3 demonstrated the capabilities of a discrete latent world model in a range of pixel-based environments \citep{hafner2023mastering}. There is growing evidence that it is easier to learn accurate models in a latent space.

Temporal abstraction has also been considered to make planning more efficient, through the use of hierarchical RL and/or options. MAXQ \cite{dietterich2000hierarchical} introduced the idea of learning hierarchical policies with multiple levels, breaking up the problem into multiple subgoals. A large literature followed, focused on efficient planning with hierarchical policies \citep{diuk2006hierarchical} and using a hierarchy of MDPs with state abstraction and macro-actions \citep{bakker2005hierarchical,konidaris2014constructing,konidaris2016constructing,gopalan2017planning}. See \citet{gopalan2017planning} for an excellent summary.

Rather than using a hierarchy and planning only in abstract MDPs, another strategy is simply to add options as additional (macro) actions in planning, still also including primitive actions. Similar ideas were explored before the introduction of options \citep{singh1992scaling, dayan1992feudal}. 
There has been some theoretical characterization of the utility of options for improving convergence rates of value iteration \citep{mann2014scaling} and sample efficiency \citep{brunskill2014pacinspired}, though also hardness results reflecting that the augmented MDP is not necessarily more efficient to solve \citep{zahavy2020planning} and hardness results around discovering options efficient for planning \citep{jinnai2019finding}. Empirically, 
incorporating options into planning has largely only been tested in tabular settings \citep{sutton1999options,singh2004intrinsically,wan2021averagereward}. Recent work has considered mechanism for identifying and learning option policies for planning under function approximation \citep{sutton2022rewardrespecting}, but as yet did not consider issues with learning the models. 

There has been some work using options for planning that is more similar to GSP, using only one-level of abstraction and restricting planning to the abstract MDP.
\citet{hauskrecht2013hierarchical} proposed to plan only in the abstract MDP with macro-actions (options) and abstract states corresponding to the boundaries of the regions spanned by the options, which is like restricting abstract states to subgoals. The most similar to our work is LAVI, which restricts value iteration to a small subset of landmark states \citep{mann2015approximate}.\footnote{A similar idea to landmark states has been considered in more classical AI approaches, under the term bi-level planning \citep{wolfe2010combined,hogg2010learning,chitnis2021learning}. These techniques are quite different from Dyna-style planning---updating values with (stochastic) dynamic programming updates---and so we do not consider them further here.} These methods also have similar flavors to using a hierarchy of MDPs, in that they focus planning in a smaller space and (mostly) avoid planning at the lowest level, obtaining significant computational speed-ups. The key distinction to GSP is that we are not in the traditional planning setting where a model is given; in our online setting, the agent needs to learn the model from interaction.

The use of landmark states has also been explored in \emph{goal-conditioned RL}, where the agent is given a desired goal state or states. This is a problem setting where the aim is to learn a policy $\pi(a| s, g)$ that can be conditioned on different possible goals. The agent learns for a given set of goals, with the assumption that at the start of each episode the goal state is explicitly given to the agent. After this training phase, the policy should generalize to previously unseen goals. Naturally, this idea has particularly been applied to navigation, having the agent learn to navigate to different states (goals) in the environment. The first work to exploit the idea of landmark states in GCRL
was for learning and using universal value function approximators (UVFAs) \citep{huang2019mapping}. The UVFA conditions action-values on both state-action pairs as well as landmark states. The agent can reach new goals by searching on a learned graph between landmark states, to identify which landmark to moves towards. A flurry of work followed, still in the goal-conditioned setting \citep{nasiriany2019planning,emmons2020sparse,zhang2020generating,zhang2021world,aubret2021distop,hoang2021successor,gieselmann2021planningaugmented,kim2021landmarkguided,dubey2021snap}. 
%Many GCRL approaches leverage UVFAs \citep{schaul2015universal}.

Some of this work focused on exploiting landmark states for planning in GCRL. \citet{huang2019mapping} used landmark states as interim subgoals, with a graph-based search to plan between these subgoals \citep{huang2019mapping}. The policy is set to reach the nearest goal (using action-values with cost-to-goal rewards of -1 per step) and learned distance functions between states and goals and between goals. These models are like our reward and discount models, but tailored to navigation and distances. \citet{nasiriany2019planning} built on this idea, introducing an algorithm called Latent Embeddings for Abstracted Planning (LEAP), that using gradient descent to search for a sequence of subgoals in a latent space. 

The idea of learning models that immediately apply to new subtasks using successor features is like GCRL, but does not explicitly use landmark states. 
The option keyboard involves encoding options (or policies) as vectors that describe the corresponding (pseudo) reward \citep{barreto2019option}. This work has been expanded more recently, using successor features \citep{barreto2020fast}. New policies can then be easily obtained for new reward functions, by linearly combining the (basis) vectors for the already learned options. However no planning is involved in that work, beyond a one-step decision-time choice amongst options.

\section{Conclusion}
In this paper we analysed a new planning framework, called Goal-Space Planning (GSP).  
GSP provides a new approach to use background planning to improve value propagation, with minimalist, local models and computationally efficient planning. 
We showed that these subgoal-conditioned models can be accurately learned using standard value estimation algorithms, and can be used to quickly propagate value through state spaces of varying sizes. We find a consequent learning speed-up on base learners with different types of value function approximation.

This work studies a new formalism, and many new technical questions along with it. We have tested GSP with pre-trained models and assumed a given set of subgoals. A critical open question is subgoal discovery. For this work, we relied on hand-chosen subgoals, but in general the agent should discover its own subgoals. In general, though, option and subgoal discovery remain open questions. One utility of this work is that it could help narrow the scope of the discovery question, to that of finding abstract subgoals that help a learner plan more efficiently.

\section*{Acknowledgments}
We thank the Alberta Machine Intelligence Institute, Alberta Innovates, the Canada CIFAR AI Chairs Program for funding this research, as well as the Digital Research Alliance of Canada for the computation resources.

\bibliography{aaai24}

\appendix
\section{Environments}
\label{app:envs}
\begin{figure}[htbp]
  \begin{centering}
  \includegraphics[width=0.2\textwidth]{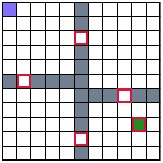}
  \caption{The FourRooms domain. The blue square is the initial state, green square the goal state, and red boxes the subgoals. A subgoal's initiation set contains the states in any room connected to that subgoal.}\label{fourrooms}  
  \end{centering}
\end{figure}

PinBall is a continuous state domain where the agent navigates a ball through a set of obstacles to reach the main goal. The environment uses a four-dimensional state representation of positions and velocities, $(x, y, \dot{x}, \dot{y}) \in [0,1] \times [0,1] \times [-2, 2] \times [-2, 2]$. The agent chooses from one of five actions at each timestep. $\mathcal{A} = \{\texttt{up}, \texttt{down}, \texttt{left}, \texttt{right}, \texttt{nothing}\}$, where the $\texttt{nothing}$ action adds no change to the ball's velocity, and the other actions each add an impulse force in one of the four cardinal directions. All collisions are elastic and we use a drag coefficient of $0.995$. This is an episodic task with a fixed starting state and main goal. An episode ends when the agent reaches the main goal or after 1,000 time steps.

\begin{figure}[h]
  \centering
  \includegraphics[width=0.2\textwidth]{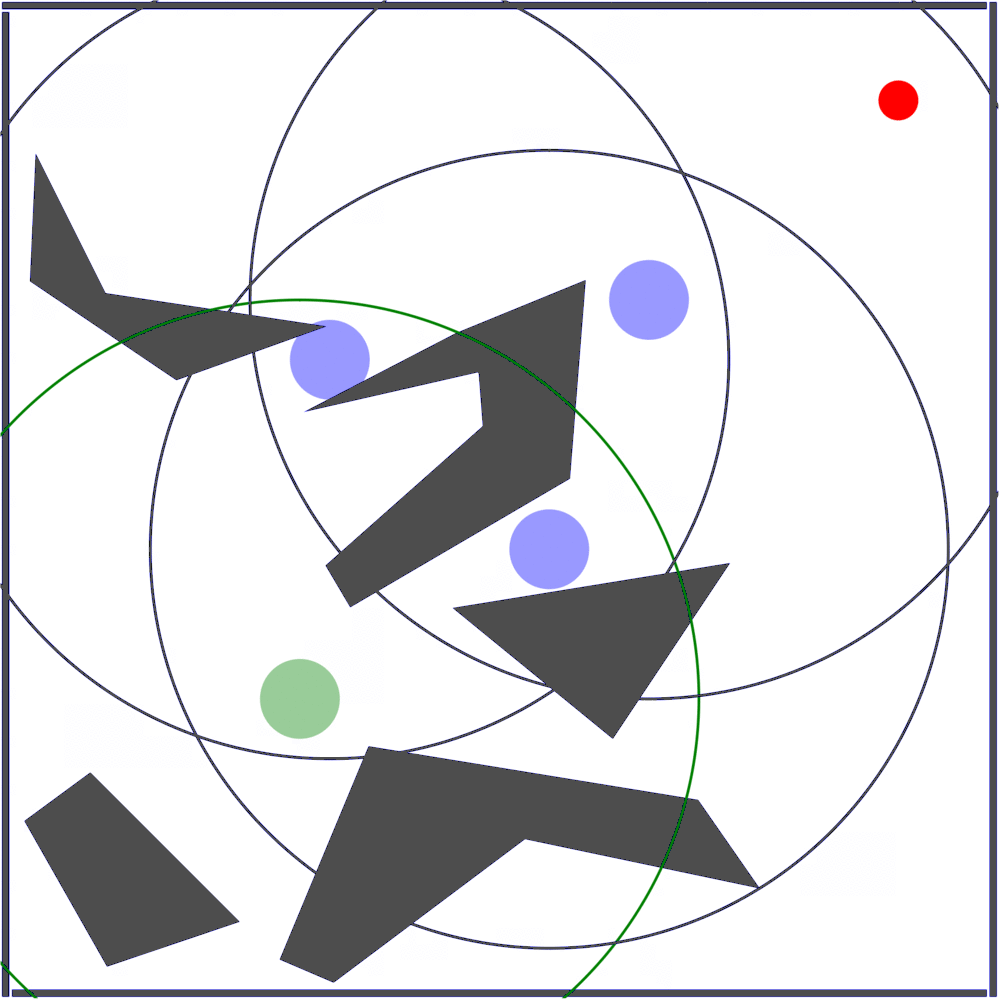}
  \caption{Obstacles and subgoals for GridBall and PinBall. The larger circles show the initiation set boundaries. Subgoals are defined in position space.}\label{fig:obstacle_layout}
\end{figure}

\section{Learning the Option Policies}
\label{app:opt_pol}
In the simplest case, it is enough to learn $\pi_g$ that makes $\rgamma(s,g)$ maximal for every relevant $s$ (i.e., $\forall \, s \in \States \, \mathrm{s.t.} \, \relsg(s,g) > 0$). For each subgoal $g$, we learn its corresponding option model $\pi_g$ by initialising the base learner in the initiation set of $g$, and terminating the episode once the learner is in a state that is a member of $g$. We used a reward of -1 per step and save the option policy once we reach a 90\% success rate, and the last 100 episodes are within some domain-dependent cut off. This cut off was 10 steps for FourRooms, and 50 steps for GridBall and PinBall.

\begin{figure*}[htbp] % You can adjust the placement options as needed
\label{fig:opt_pol_eval}
 \centering
    \includegraphics[width=\textwidth]{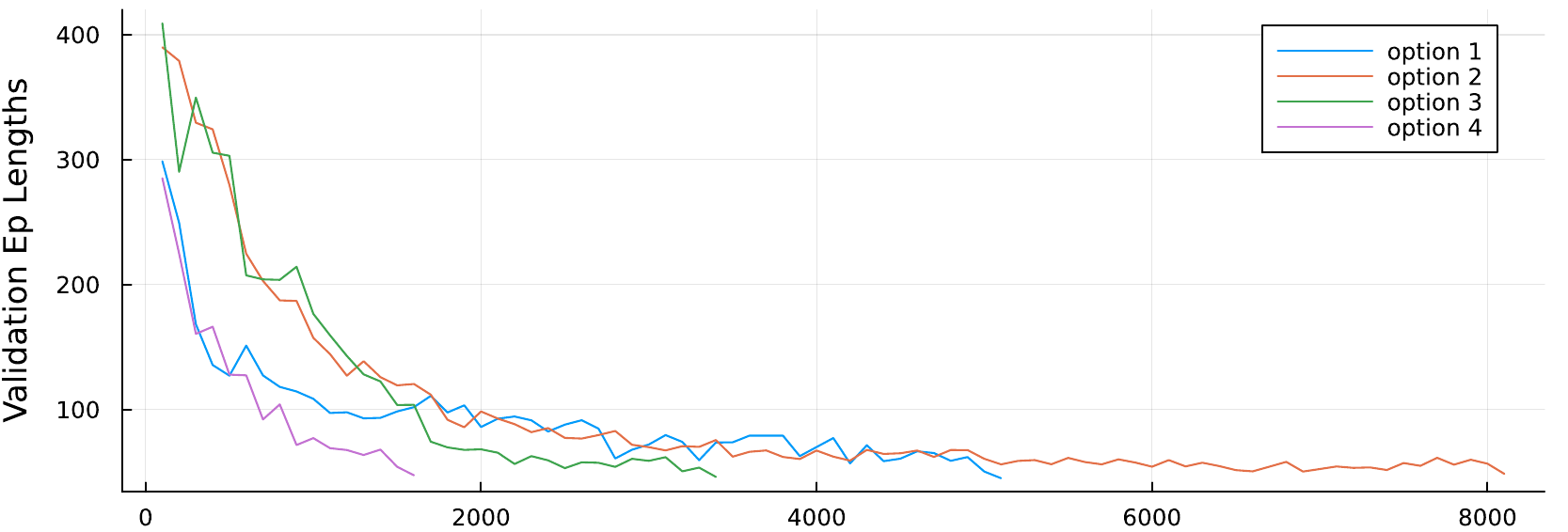}
    \caption{Evaluation of PinBall option policies by average trajectory length. Policies were saved once they were able to reach their respective subgoal in under 50 steps, averaged across 100 trajectories. Subgoal 2 was the hardest to learn an option policy for, due to its proximity to obstacles.}
\end{figure*}

\textbf{Hyperparameters} In FourRooms, we use Sarsa(0) and Sarsa(0.9) base learners with learning rate $\alpha = 0.01$, discount factor $\gamma_c = 0.99$ and an $\epsilon = 0.02$ for its $\epsilon$-greedy policy. In GridBall, we used Sarsa(0) and Sarsa(0.9) base learners with $\alpha=0.05$, $\gamma_c=0.99$ and $\epsilon=0.1$. $\epsilon$ is decayed by 0.5\% each timestep. In the linear function approximation setting, these learners use a tilecoder with 16 tiles and 4 tilings across each of the both the GridBall dimensions. In PinBall, the Sarsa(0.9) learner was tuned to $\alpha = 0.1$, $\gamma_c=0.99$, $\epsilon=0.1$, decayed in the same manner as in GridBall. The same tile coder was used on on the 4-dimensional state space of PinBall. For the DDQN base learner, we use $\alpha=0.004$, $\gamma_c=0.99$, $\epsilon=0.1$, a buffer that holds up to $10,000$ transitions a batch size of $32$, and a target refresh rate of every $100$ steps. The Q-Network weights used Kaiming initialisation \citep{he2015delving}.

We could have also learned the action-value variant $\rgamma(s,a,g)$ using a Sarsa update, and set $\pi_g(s) = \argmax_{a \in \Actions} \rgamma(s,a,g)$, where we overloaded the definition of $\rgamma$. We can then extract $\rgamma(s,g) = \max_{a \in \Actions} \rgamma(s,a,g)$, to use in all the above updates and in planning. In our experiments, this strategy is sufficient for learning $\pi_g$.

\section{Learning the Subgoal Models}
\label{app:model_learning}
In our experiments, the data is generated offline according to each $\pi_g$. We then use this episode dataset from each $\pi_g$ to learn the subgoal models for that subgoal $g$. This is done by ordinary least squares regression to fit a linear model in four-room, and by stochastic gradient descent with neural network models in GridBall and PinBall. 

We first collect a dataset of $n$ episodes leading to a subgoal $g$,  $\mathcal{D}_g = \{\langle S_{i,1}, A_{i,1}, R_{i,1}, S_{i,1}, \dots, S_{i,T_i} \rangle\}_{i=1}^{n}$. $S_{i,t}, A_{i,t}, R_{i, t}$ represent the state, action and reward at timestep $t$ of episode $i$. $T_i$ is the length of episode $i$. $S_{i,0}$ is a randomised starting state within the initiation set of $g$, and $S_{i,T_i}$ is a state that is a member of subgoal $g$. For each $g$, we use $\mathcal{D}_g $ to generate a matrix of all visited states, $\mathbf{X} \in \mathbb{R}^{l\times|\States|}$, and a vector of all reward model returns, $\mathbf{g}_r \in \mathbb{R}^l$, and transition model returns $\mathbf{g}_\gamma \in \mathbb{R}^l$,
\begin{align*}
\mathbf{X} = \begin{pmatrix}
              S_{i,1} \\
              S_{i,2} \\
              \vdots \\
              S_{n,T_{n}}
              \end{pmatrix},
\mathbf{g}_r = \begin{pmatrix}
              R_{i,2} + \gamma \rgamma(S_{i,3}, g) \\
              R_{i,3} + \gamma \rgamma(S_{i,4}, g)\\
              \vdots \\
              R_{n,T_n}
              \end{pmatrix},\\
\mathbf{g}_\gamma = \begin{pmatrix}
                \gamma^{T_1 - 0} \\
                \gamma^{T_1 - 1} \\
                \vdots \\
                \gamma^{T_{n} - T_{n}}
                \end{pmatrix},\\
\end{align*}
where $l = \sum_{i = 1}^n T_i$ is the total number of visited states in $\mathcal{D}_g$.

This creates a system of linear equations, whose weights we can solve for numerically in the four-room domain,
\begin{align*}
  \mathbf{X}\rparams = \mathbf{g}_r \implies \rparams = \mathbf{X}^+\mathbf{g}_r, \\
  \mathbf{X}\gamparams = \mathbf{g}_\gamma \implies \gamparams = \mathbf{X}^+\mathbf{g}_\gamma,
\end{align*}
where $\rparams, \gamparams \in \mathbb{R}^{|\mathcal{S}|}$ and $\mathbf{X}^+$ is the Moore-Penrose pseudoinverse of $\mathbf{X}$ \citep{penrose1955generalized}.

For GridBall and PinBall, we used fully connected artificial neural networks for $\rgamma$ and $\Gamma$, and performed mini-batch stochastic gradient descent to solve $\rparams$ and $\gamparams$ for that subgoal $g$. We use each mini-batch of $m$ states, reward model returns and transition model returns to perform the update:
\begin{align*}
  \rparams \gets\rparams - \eta_r \sum_{j=1}^m \nabla_\rparams(\rparams^\top\mathbf{X}_{j,:} - \mathbf{g}_{r,j})^2,\\
  \gamparams \gets\gamparams - \eta_\Gamma \sum_{j=1}^m \nabla_\gamparams(\gamparams^\top\mathbf{X}_{j,:} - \mathbf{g}_{\gamma,j})^2,
\end{align*}
where $\eta_r$ and $\eta_\Gamma$ are the learning rates for the reward and discount models respectively. $\mathbf{X}_{j,:}$ is the $j^{\mathrm{th}}$ row of $\mathbf{X}$. $\mathbf{g}_{r,j}$ and $\mathbf{g}_{\gamma,j}$ are the $j^{\mathrm{th}}$ entry of $\mathbf{g}_{r}$ and $\mathbf{g}_{\gamma}$ respectively. In our experiments, we had a fully connected artificial neural network with two hidden layers of 128 units and ReLU activation for each subgoal. The network took a state $s = (x, y, \dot{x}, \dot{y})$ as input and outputted both $\rgamma(s,g)$ and $\Gamma(s,g)$. All weights were initialised using Kaiming initialisation \citep{he2015delving}. We use the Adam optimizer with $\eta=0.001$ and the other parameters set to the default ($b_1=0.9, b_2=0.999, \epsilon=10^{-8}$), mini-batches of 1024 transitions and 100 epochs.

\begin{figure}[htbp]
  \centering
  \begin{minipage}[b]{0.45\textwidth}
      \centering
      \caption*{$\rgamma(s,g_1)$ and $\GamModel(s, g_1)$}
      \includegraphics[width=\textwidth]{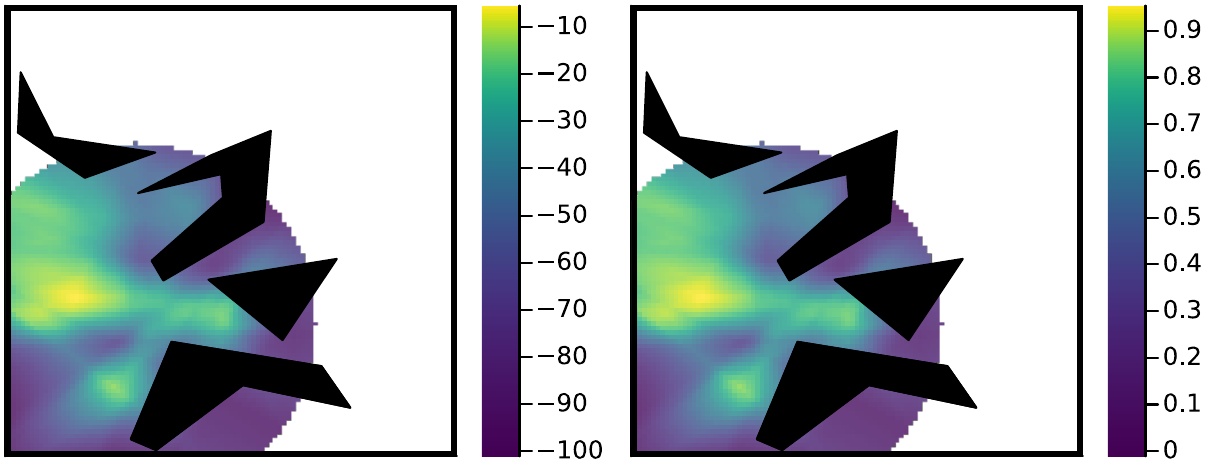}
  \end{minipage}
  \begin{minipage}[b]{0.45\textwidth}
      \centering
      \caption*{$\rgamma(s,g_2)$ and $\GamModel(s, g_2)$}
      \includegraphics[width=\textwidth]{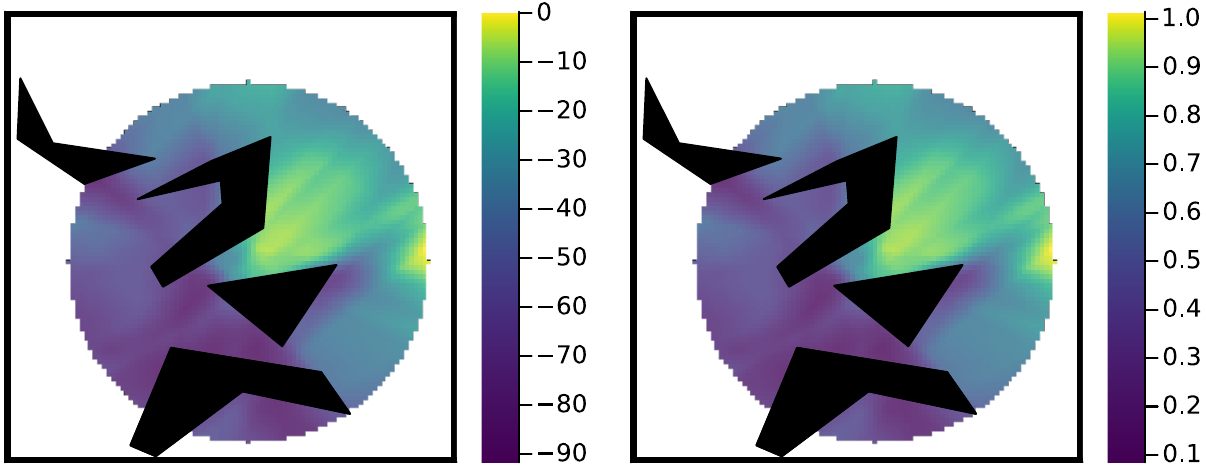}
  \end{minipage}\newline
  \begin{minipage}[b]{0.45\textwidth}
      \centering
      \caption*{$\rgamma(s,g_3)$ and $\GamModel(s, g_3)$}
      \includegraphics[width=\textwidth]{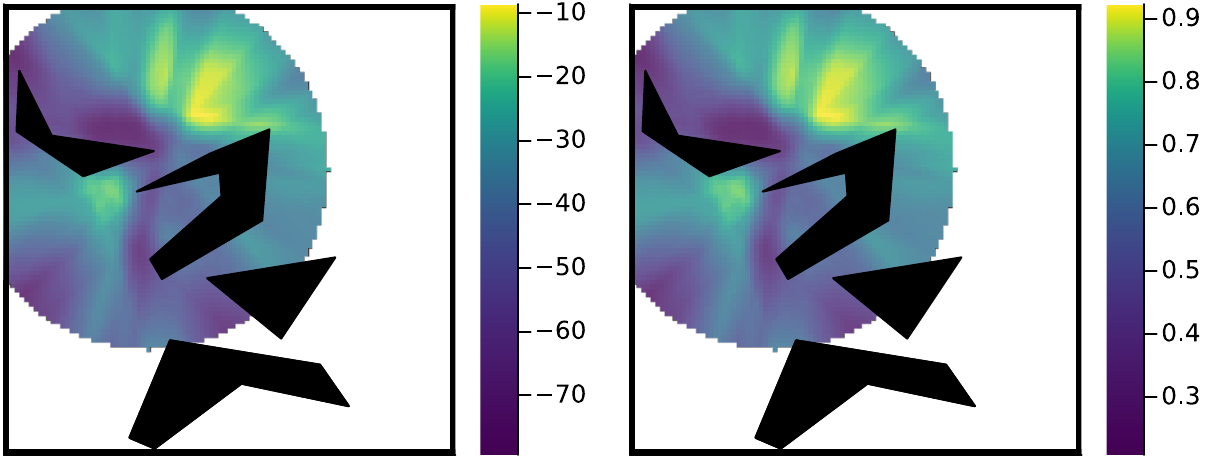}
  \end{minipage}
  \begin{minipage}[b]{0.45\textwidth}
      \centering
      \caption*{$\rgamma(s,g_4)$ and $\GamModel(s, g_4)$}
      \includegraphics[width=\textwidth]{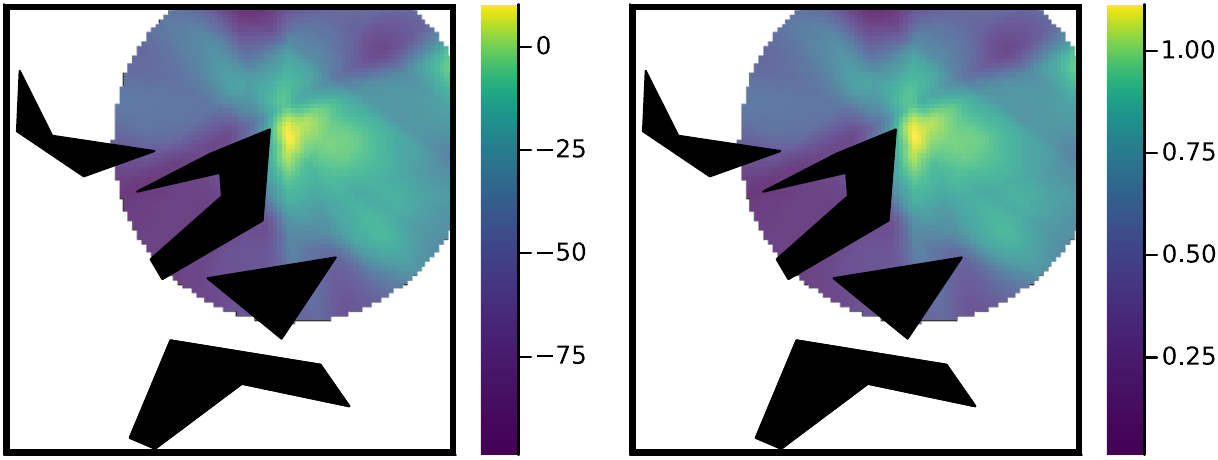}
  \end{minipage}
\caption{State-to-Subgoal models learnt by neural models after 100 epochs.}
\label{fig:s2g_models}
\end{figure}

\section{Pseudocode}

\begin{algorithm}[H]
  \caption{\subroutine{MainPolicyUpdate}$(s, a, r, s', \gamma, a')$}
  \label{alg:MainPolicySarsaLambdaUpdate}
\begin{algorithmic}
\STATE // For a Sarsa($\lambda$) base learner
\STATE  $\vsub \gets \max_{g \in \augGoals: \relsg(s,g) > 0} \rgamma(s, g; \mparams) + \GamModel(s,g; \mparams) \vgoal(g)$
  \STATE $\delta \leftarrow r + \gamma \vsub(s') - \vsub(s) + \gamma q(s',a'; \qparams) - q(s,a; \qparams)$
  \STATE $\qparams \leftarrow \qparams + \alpha \delta \mathbf{z}\nabla_\qparams q(s,a; \qparams)$ 
  \STATE $\mathbf{z} \gets \gamma\lambda\mathbf{z} + \nabla_\qparams q(s,a; \qparams)$
\end{algorithmic}
\end{algorithm}

\begin{algorithm}[H]
  \caption{\subroutine{Planning}$()$}
  \label{alg:GoalSpacePlanning}
\begin{algorithmic}
  \FOR {$n$ iterations, for each $g \in \Goals$ }
    \STATE $\vgoal(g) \leftarrow \underset{g' \in \augGoals: d(g, g') > 0}{\max} \vgg(g, g'; \vggparams) + \Gammagg(g, g'; \gamggparams) \vgoal(g')$
  \ENDFOR
\end{algorithmic}
\end{algorithm}

\begin{algorithm}[H]
  \caption{\subroutine{ModelUpdate}$(s, a, r, s', \gamma)$}
  \label{alg:ModelUpdate}
\begin{algorithmic}
\STATE Add new transition $(s, a, s', r, \gamma)$ to buffer $B$
  \FOR {$g' \in \augGoals$, for multiple transitions $(s, a, r, s', \gamma)$ sampled from $B$}
  \STATE $\gamma_{g'} \gets \gamma (1 - \indsg(s', g'))$ 
  \STATE // Update option policy - e.g. by Sarsa
   \STATE $a' \gets \pi_{g'}(s'; \polparams)$   
    \STATE $\delta^\pi \leftarrow \tfrac{1}{2} (r - 1) + \gamma_{g'} \optionq(s', a', g'; \polparams) - \optionq(s, a, g'; \polparams)$
    \STATE $\polparams \leftarrow \polparams + \alpha^{\pi}\delta^\pi \nabla_\polparams q(s, a, g'; \polparams)$
    \STATE // Update reward model and discount model
 % \State $a'_{\text{max}} \gets \argmax_{a' \in \Actions} q(s', a', g'; \theta^\pi)$ (greedy action for option policy)
    \STATE $\delta^r \leftarrow r + \gamma_{g'} \rgamma(s', a', g'; \rparams) - \rgamma(s, a, g'; \rparams)$
    \STATE $\delta^\Gamma \leftarrow \indsg(s', g)\gamma + \gamma_{g'} \GamModel(s', a', g'; \gamparams) - \GamModel(s, a, g'; \gamparams)$
    \STATE $\rparams \leftarrow \rparams + \alpha^{r}\delta^r \nabla_\rparams \rgamma(s, a, g'; \rparams)$
    \STATE $\gamparams \leftarrow \gamparams + \alpha^{\Gamma} \delta^\Gamma \nabla_\gamparams \GamModel(s, a, g'; \gamparams)$
    \STATE // Update goal-to-goal models using state-to-goal models
%    \State $a_{\text{max}} \gets \argmax_{a \in \Actions} q(s, a, g'; \theta^\pi)$
    \FOR {each $g$ such that $\indsg(s,g) > 0$ }
    \STATE $\vggparams \leftarrow \vggparams + \tilde{\alpha}^{r} (\rgamma(s, g'; \mparams) - \vgg(g, g';\vggparams)) \nabla_\rparams \vgg(g, g'; \vggparams)$
     \STATE $\gamggparams \leftarrow \gamggparams + \tilde{\alpha}^{\Gamma} (\GamModel(s, g'; \mparams) - \Gammagg(g, g';\vggparams)) \nabla_\gamparams
     \Gammagg(g, g'; \gamggparams)$
 \ENDFOR
  \ENDFOR
\end{algorithmic}
\end{algorithm}

It is simple to extend the above pseudocode for the main policy update and the option policy update to use Double DQN \citep{van2016deep} updates with neural networks. The changes from the above pseudocode are 1) the use of a target network to stabilize learning with neural networks, 2) changing the one-step bootstrap target to the DDQN equivalent, 3) adding a replay buffer for learning the main policy, and 4) changing the update from using a single sample to using a batch update. Because the number of subgoals is discrete, the equations for learning $\vggparams$ and $\gamggparams$ does not change. We previously summarized these changes for learning the main policy in Algorithm \ref{alg:DDQN_GSP} and now detail the subgoal model learning in Algorithm \ref{alg:DDQN_model}.

\begin{algorithm}[htbp]
\caption{\subroutine{{\color{blue}GSP} (built on DDQN)}}
\label{alg:DDQN_GSP}
\begin{algorithmic}
\STATE Initialize base learner parameters $\qparams,\qtarg = \qparams_0$, {\color{blue} set of subgoals $\Goals$, relevance function $d$}

\STATE Sample initial state $s_0$ from the environment
  \FOR {$t \in 0, 1, 2, ...$}
      \STATE Take action $a_t$ using $q$ (e.g., $\epsilon$-greedy),
      \STATE Observe $s_{t+1}, r_{t+1}, \gamma_{t+1}$
  \STATE Add $(s_t, a_t, s_{t+1}, r_{t+1}, \gamma_{t+1})$ to replay buffer $D$    
  \STATE {\color{blue}\subroutine{DDQNModelUpdate}$()$ (see Algorithm \ref{alg:DDQN_model})}
  \STATE {\color{blue}\subroutine{Planning}$()$ (see Algorithm \ref{alg:GoalSpacePlanning})}
  \FOR{$n$ mini-batches}
    \STATE Sample batch $B = \{ (s, a, r, s', \gamma )\}$ from $D$
    {\color{blue} \IF{$d(s, \cdot), d(s', \cdot) > 0$}
      \STATE $\vsub(s) = \underset{g \in \augGoals: \relsg(s,g) > 0}{\mathrm{max}}\, \rgamma(s, g) + \GamModel(s,g) \vgoal(g)$
      \STATE $\vsub(s') = \underset{g \in \augGoals: \relsg(s',g) > 0}{\mathrm{max}} \rgamma(s', g) + \GamModel(s',g) \vgoal(g)$
      \STATE $\tilde{r} = r + \gamma \vsub(s') - \vsub(s) $
    \ELSE
      \STATE $\tilde{r} = r$
    \ENDIF
    \STATE $Y = \tilde{r} + \gamma q(s',\argmax_{a'} q(s',a'; \qparams); \qtarg)$}
    \STATE $L = \frac{1}{|B|}\underset{_{(s, a, r, s', \gamma) \in B}}{\sum} (Y(s,a,r,s',\gamma) - q(s,a; \qparams))^2$
    \STATE $\qparams \leftarrow \qparams - \alpha \nabla_\qparams L$ 
    \IF{$n_{\text{updates}}\% \tau == 0$}
    \STATE $\qtarg \leftarrow \qparams $
    \ENDIF
    \STATE $n_{\text{updates}}$ = $n_{\text{updates}}$ + 1
  \ENDFOR
  \ENDFOR
\end{algorithmic}
\end{algorithm}

\begin{algorithm}[htbp]
  \caption{\subroutine{DDQNModelUpdate}$(s, a, r, s', \gamma)$}
  \label{alg:DDQN_model}
\begin{algorithmic}
\STATE Add new transition $(s, a, s', r, \gamma)$ to buffer $D_{\mathrm{model}}$
  \FOR {$g' \in \augGoals$}
  \FOR {$n_{\mathrm{model}}$ mini-batches} 
  \STATE Sample batch $B_{\mathrm{model}} = \{ (s, a, r, s', \gamma )\}$ from $D_{\mathrm{model}}$
    % \For {for multiple transitions $(s, a, r, s', \gamma)$ sampled from $B$}
  \STATE $\gamma_{g'} \gets \gamma (1 - \indsg(s', g'))$ 
  \STATE // Update option policy
  \STATE $a' \leftarrow \argmax_{a' \in \Actions} \optionq(s', a', g'; \polparams)$
    \STATE $\delta^\pi(s, a, s', r, \gamma) \gets \tfrac{1}{2} (r - 1) + \gamma_{g'} \optionq(s', a', g'; \boldsymbol{\theta}^\pi_{\mathrm{targ}}) - q(s, a, g'; \polparams)$
    \STATE $\polparams \leftarrow \polparams - \alpha^{\pi}\nabla_{\polparams} \frac{1}{|B_{\mathrm{model}}|} \sum_{(s, a, r, s', \gamma) \in B_{\mathrm{model}}}(\delta^\pi)^2$

    \STATE $\polparams_{\mathrm{targ}} \leftarrow \rho_{\mathrm{model}} \polparams + (1 - \rho_{\mathrm{model}})\polparams_{\mathrm{targ}}$ 
    \STATE // Update reward model and discount model
 % \State $a'_{\text{max}} \gets \argmax_{a' \in \Actions} q(s', a', g'; \theta^\pi)$ (greedy action for option policy)
    \STATE $\delta^r \leftarrow r + \gamma_{g'}(\gamma, s')\rgamma(s', a', g'; \rparams_{\mathrm{targ}}) - \rgamma(s, a, g'; \rparams)$
    \STATE $\delta^\Gamma  \leftarrow \indsg(s', g)\gamma + \gamma_{g'}(\gamma, s') \GamModel(s', a', g'; \gamparams_{\mathrm{targ}}) - \GamModel(s, a, g'; \gamparams)$
    \STATE $\rparams \leftarrow \rparams - \alpha^{r} \nabla_{\rparams} \frac{1}{|B_{\mathrm{model}}|} \underset{_{(s, a, r, s', \gamma) \in B}}{\sum}(\delta^r)^2$
    \STATE $\gamparams \leftarrow \gamparams - \alpha^{\Gamma} \nabla_{\gamparams} \frac{1}{|B_{\mathrm{model}}|} \underset{_{(s, a, r, s', \gamma) \in B}}{\sum} (\delta^\Gamma)^2$
    \IF{$n_{\text{updates}}\% \tau == 0$}
    \STATE $\rparams_{\mathrm{targ}} \leftarrow \rparams$ 
    \STATE $\gamparams_{\mathrm{targ}} \leftarrow \gamparams$
    \ENDIF
    \STATE $n_{\text{updates}}$ = $n_{\text{updates}}$ + 1
    \ENDFOR
    \STATE // Update goal-to-goal models using state-to-goal models
    \STATE \dots same as in prior pseudocode.
  \ENDFOR
\end{algorithmic}
\end{algorithm}

\end{document}